\documentclass{article}
\usepackage{spconf,amsmath}
\usepackage{comment}
\usepackage{placeins}
\usepackage{graphicx}
\usepackage{booktabs}
\usepackage{caption}
\usepackage{subcaption}

\usepackage[utf8]{inputenc} 
\usepackage[T1]{fontenc}    
\usepackage{hyperref}
\hypersetup{
    colorlinks=true,
    linkcolor=blue,
    urlcolor=blue}
\usepackage{url}            
\usepackage{booktabs}       
\usepackage{amsfonts}       
\usepackage{nicefrac}       
\usepackage{microtype}      
\usepackage{xcolor}

\usepackage{todonotes}

\title{HUMBUG ZOONIVERSE: A CROWD-SOURCED ACOUSTIC MOSQUITO DATASET}
\name{\begin{tabular}{c} Ivan Kiskin$^{\star}$ \qquad Adam D. Cobb$^{\star}$ \qquad Lawrence Wang$^{\star}$ \qquad Yunpeng Li$^{\star\star}$ \qquad Davide Zilli$^{\ddagger}$ \\
Marianne Sinka$^{\star\star\star}$ \qquad Kathy Willis$^{\star\star\star}$ \qquad Stephen Roberts$^{\star \ddagger}$
\end{tabular}}
\address{$^{\star}$Department of Engineering Science, University of Oxford\\
$^{\star\star}$Department of Computer Science, University of Surrey\\
$^{\star\star\star}$Department of Zoology, University of Oxford\\
        $^{\ddagger}$Mind Foundry Ltd.
        }

\begin{document}
%
\maketitle

\begin{abstract}
Mosquitoes are the only known vector of malaria, which leads to hundreds of thousands of deaths each year. Understanding the number and location of potential mosquito vectors is of paramount importance to aid the reduction of malaria transmission cases. In recent years, deep learning has become widely used for bioacoustic classification tasks. In order to enable further research applications in this field, we release a new dataset of mosquito audio recordings. With over a thousand contributors, we obtained 195,434 labels of two second duration, of which approximately 10 percent signify mosquito events. We present an example use of the dataset, in which we train a convolutional neural network on log-Mel features, showcasing the information content of the labels.  We hope this will become a vital resource for those researching all aspects of malaria, and add to the existing audio datasets for bioacoustic detection and signal processing.

\end{abstract}

\section{Introduction}



Malaria is one of the most severe public health problems in the developing world. The World Health Organization estimated that there were 219 million malaria cases worldwide in 2017, leading to 435,000 related deaths \cite{WHO2019}. Vector control efforts have achieved significant improvement in the past few decades \cite{bhatt2015effect}. However, the effect of malaria intervention strategies remains poorly understood due to the absence of reliable surveillance data. 

Malaria is transmitted through the bite of an infected Anopheles mosquito. Among the approximately 3,600 species of mosquitoes, only about 60 out of the 450 Anopheles species transmit malaria (i.e., are vectors) \cite{Neafsey2015, wilkerson2015making}. The ability to quickly detect the presence of these mosquito species is therefore crucial for control programmes and targeted intervention strategies.

In recent years, the process of mosquito (alongside other insect) detection has been undergoing automation, with a range of applications of neural networks in the domain of audio data \cite{kiskin2018bioacoustic,PotamitisKaggle}. To aid the potency of existing algorithms and encourage the development of more data-driven approaches, we release this crowd-sourced dataset. Our dataset contains a mixture of mosquito species recorded in both laboratory and field conditions (the detailed distribution is given in Section \ref{sec:data_acquistion}). We showcase the metadata, and present an application of a simple, yet effective, convolutional neural network trained on log-Mel \cite{hershey2017cnn} spectrogram features. This represents the basis of many neural networks currently used in this problem domain \cite{deng2014deep, langkvist2014review}. Furthermore, we suggest methods for dealing with difficulty of this real-world data, namely ways of dealing with data imbalance and label coverage.

The remainder of the paper is structured as follows. In Section \ref{sec:priorwork}, we place our work in the context of prior work. In Section \ref{sec:data} we give details of the dataset, with meta-information of the distribution of data labels present within the recording. Section \ref{sec:baseline} shows a simple implementation of a convolutional neural network to demonstrate the detectability of events on random splits of the data. Finally, we conclude in Section \ref{sec:conclusion}.

\begin{figure*}[t]
    \centering
    \includegraphics[trim=0 0 0 0,clip,width=0.8\textwidth]{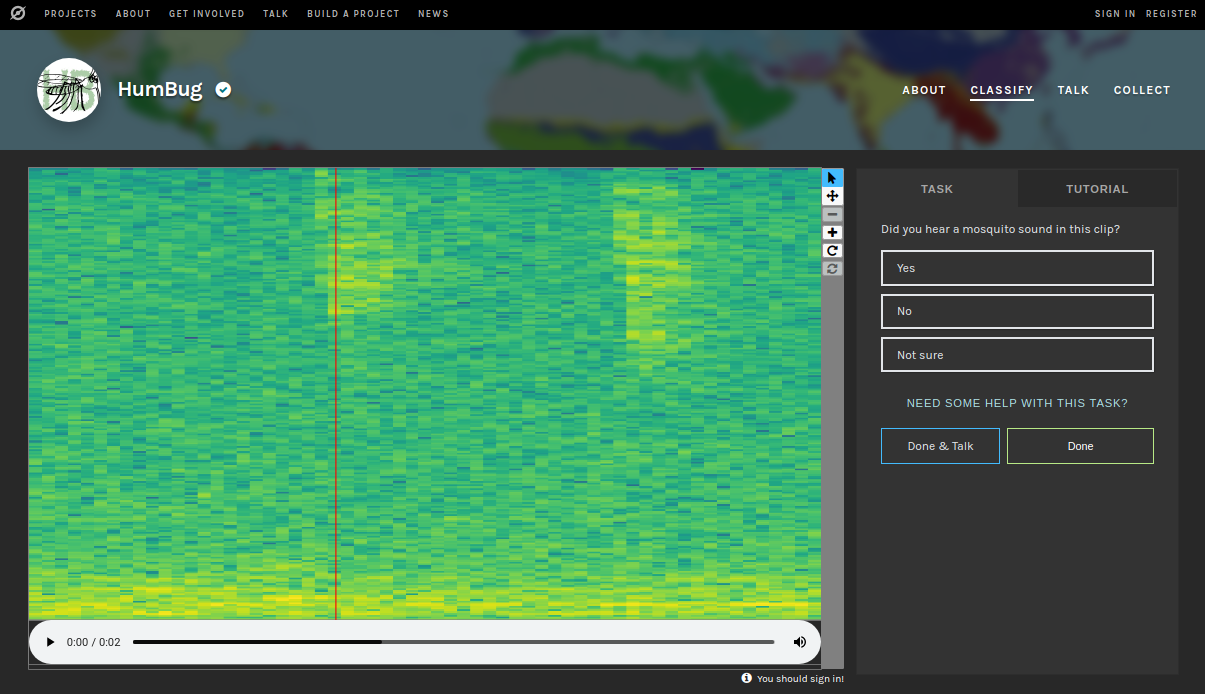} 
    \caption{User interface for the classification page of the Zooniverse HumBug project, found at \url{https://www.zooniverse.org/projects/yli/humbug/}. A short-time Fourier transform spectrogram representation is offered above the audio file.}
    \label{fig:zooniverseUI}
\end{figure*}



\begin{figure*}[h!]
     \centering
     \begin{subfigure}[t]{1.\columnwidth}
        ~~~~~~~~
        \includegraphics[trim=0 30 485 0,clip,width=.7\textwidth]{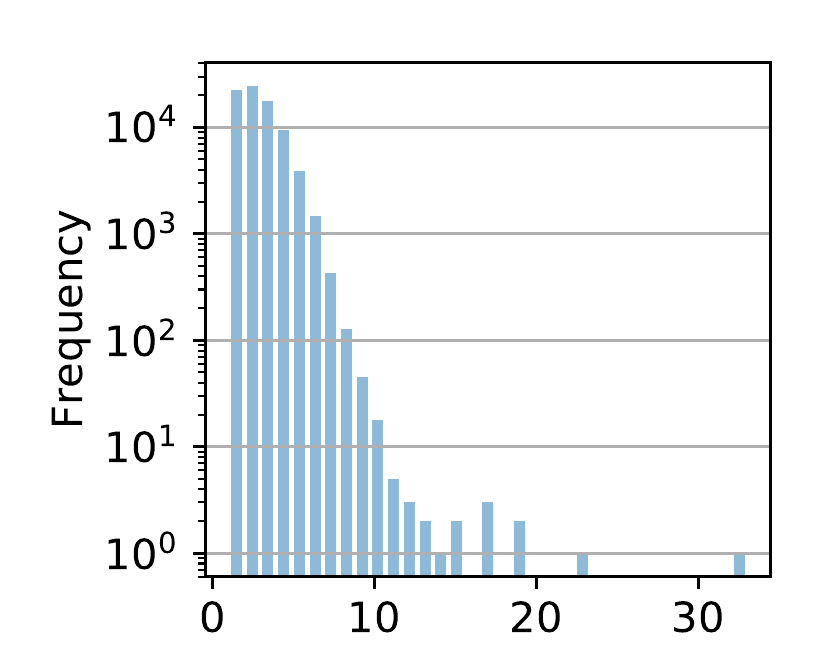}
        \caption{Labels per clip}
         \label{fig:labelsperclip}
     \end{subfigure}%
            \begin{subfigure}[t]{1.\columnwidth}
        ~~~~~~~~    
  \includegraphics[trim=235 30 250 0,clip,width=.7\textwidth]{Imagesv2/Metadata_not_sure_into_yes_single_not_sure_into_0_5_multi_yes_overlap_smalltext.pdf}    \caption{User number}
         \label{fig:labelsperuser}
     \end{subfigure}
     
       \begin{subfigure}[t]{1.\textwidth}
         \centering
        \includegraphics[trim=0 0 0 0,clip,width=0.9\textwidth]{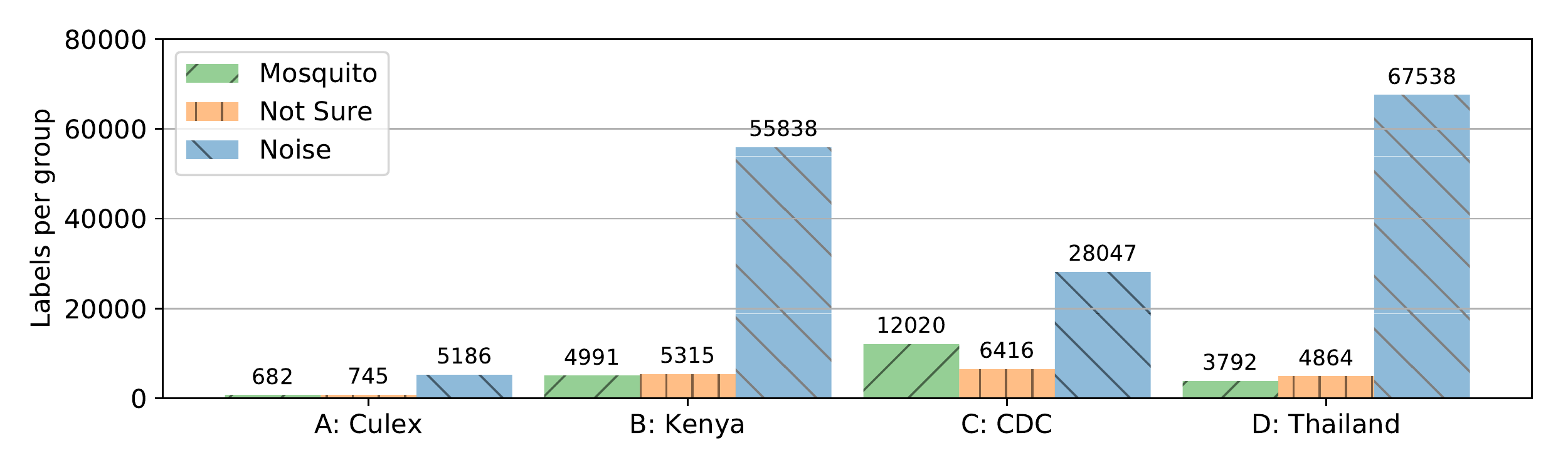} 
    \caption{Responses of raw votes available in the database, split by recording group}
         \label{fig:subject_set}
     \end{subfigure}
    \caption{Statistics of the crowd-sourced Zooniverse dataset. The dataset is comprised of four sources of data, labelled ``\textbf{A}'', ``\textbf{B}'', ``\textbf{C}'', and ``\textbf{D}'', described in Section \ref{sec:data_acquistion}.  The total number of classifications made is 195,434. This is made on 80,101 overlapping 2 second chunks (each with a unique $\mathtt{audio\_id}$, which create a 22 hour dataset of unique audio. 57,710 (72\%) of the audio samples contain more than one label, of which 10,487 (18\%) contain disagreement. In total, approximately 1 in 10 recordings contains an audible mosquito, with the distribution given in (c).}
    \label{fig:metadata}
\end{figure*}

\begin{figure*}
     \centering
     \begin{subfigure}[t]{0.47\textwidth}
         \centering
        \includegraphics[trim=0 0 0 0,clip,width=1.0\textwidth]{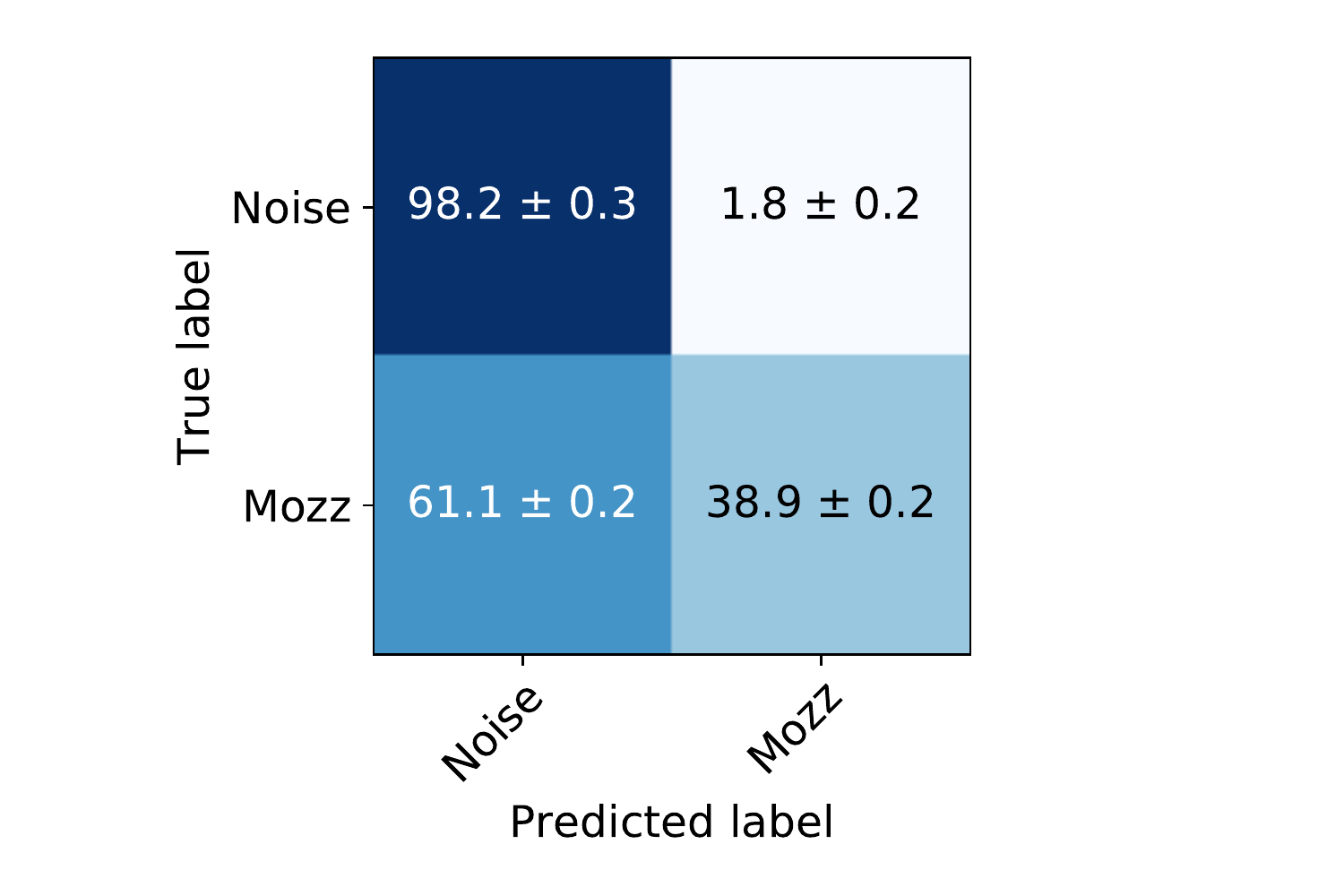} 
        \caption{Cross-entropy class weights $w_0 = 1, w_1 = 1$}
         \label{fig:confmata}
     \end{subfigure}
     \begin{subfigure}[t]{0.47\textwidth}
         \centering
        \includegraphics[width=1.0\textwidth]{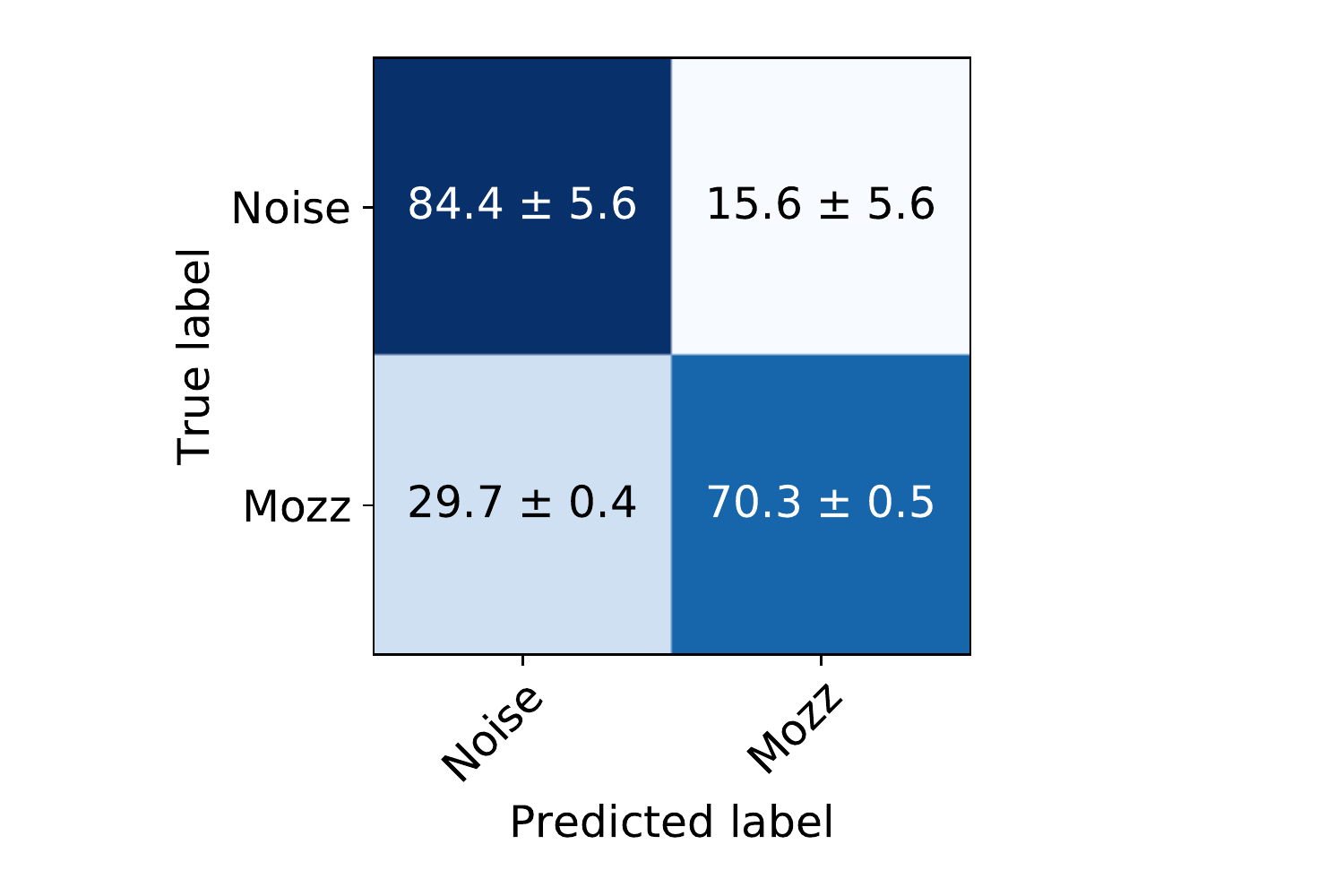} 
    \caption{Cross-entropy class weights $w_0 = 1, w_1 = 10$}
         \label{fig:confmatb}
     \end{subfigure}
    \caption{Mean confusion matrices of classifications made with the CNN baseline on ten folds of the dataset, given in the form of $\mu \pm \sigma$, where $\mu$ is the mean, and $\sigma$ is the standard deviation. The two subfigures show two weighted cross-entropies which encode our utility in detecting either class. The trade-off of false negative and true positive rate is evidenced when we change the weighting of each class in the loss function.}
    \label{fig:confmat}
\end{figure*}

\section{Relation to Prior work}
\label{sec:priorwork}
The identification of mosquitoes by their flight tones has been researched for more than half a century \cite{Khan1949}.  In recent years there has been a focus on employing machine learning techniques to aid with detection, which has seen some success \cite{kiskin2018bioacoustic,PotamitisKaggle}.
To further research in this area, we build a dataset which can help with the development of more reliable detection algorithms.
One example of an already-existing dataset is the ``Wingbeats'' pseudo-acoustic data, which consists of 279,566 short recordings (of approximately one second duration) \cite{PotamitisKaggle}. One potential disadvantage of this data is that it is collected via optical sensors, which may not translate well to detection with low-cost acoustic sensors. In order to provide a new dataset that may be more suited to helping with the detection of mosquitoes in the developing world, our data is collected via conventional microphones that are found in low-cost mobile phones. Collecting via these microphones allow both greater participation from those who already live in malaria-ridden areas, and the possibility of providing a method for detection in people's homes. Specifically, these smartphones can be placed in bed net corners (a form of intervention), where mosquitoes tend to congregate, to allow the autonomous collection of data. This is a suggested strategy currently under research as part of the HumBug project\footnote{\url{humbug.ac.uk}}. There has already been research that uses similar technology \cite{li2017mosquito,eLifeArticle} and we believe that this dataset will only help further to solve these kinds of issues.

Furthermore, we highlight the utility of crowd-sourcing labels, where we have gathered labels for our data via the Zooniverse platform. The Zooniverse platform is a useful tool for labelling data (see Figure \ref{fig:zooniverseUI}) and has seen success in multiple other projects that help with machine learning for the developing world, such as disaster response and earthquake detection \cite{Simpson2014}.


\section{Data}
\label{sec:data}
We supply the label information in a $\mathtt{csv}$ file with the following fields: $\mathtt{\{audio\_id, id, labeller\_id, zooniverse\_id},$ $  \mathtt{sound\_type, path, subject\_set}\}$. The $\mathtt{path}$ field links the label to the appropriate wave file. The class label is contained in $\mathtt{sound\_type}$, and each labeller is assigned a unique $\mathtt{labeller\_id}$. This leaves the practitioner free to choose whichever voting scheme desirable. The batch that describes the experiment the data was collected from is given in $\mathtt{subject\_set}$. This is included for considerations of recording devices, which may vary across the different data collection procedures described in Section \ref{sec:data_acquistion}.\footnote{See the \href{https://github.com/HumBug-Mosquito/ZooniverseData}{GitHub}  repository for the $\mathtt{csv}$ file and corresponding $\mathtt{wav}$ audio data.}

\subsection{Data acquisition}
\label{sec:data_acquistion}
The recordings presented are from 4 sources, and indexed by $\mathtt{subject\_set}$ according to the following:

\textbf{Group A} consists of laboratory-based mosquito colonies held in the UK (Oxford), providing acoustic data for vector species Culex quinquefasciatus, a vector of the West Nile virus. These culture cages contained both male and female insects. 

\textbf{Group B} was acquired form laboratory-based mosquito colonies in the United States Army Medical Research Unit, Kenya (USAMRU-K), providing acoustic data for Anopheles gambiae, the primary vector of malaria in Africa. 

\textbf{Group C} was made with the recording of multiple species at the insectary at the Center for Disease Control and Prevention (CDC), Atlanta, USA, including Aedes aegypti and Aedes albopictus, vectors of yellow and dengue fever respectively. 

\textbf{Group D} is formed from recordings of wild captured mosquitoes (including Anopheles barbirostris and Anopheles maculatus, Asian vectors of malaria), sampled form the Pu Teuy Village, Sai Yok District, Kanchanaburi Province, Thailand. This site is a field site of Kaetsart University, Bangkok.

 Figure \ref{fig:subject_set} shows the proportion of data labelled as each class within the four sources of recordings which constitute the whole dataset. We note groups \textbf{A, B} and \textbf{D} show a similar fraction (approx 10\%) of mosquito events, with group \textbf{C} containing the highest relative frequency of clearly audible mosquitoes.

\subsection{Data labelling}
With large-scale field deployment, the number of recordings requiring data tagging is beyond the capacity of experts and researchers in the HumBug project. We hence resorted to the power of crowdsourcing, creating a project on Zooniverse, the world’s largest citizen science platform \cite{Simpson2014}, to solicit labels from over a million volunteers. Volunteers listen to 2 second sound clips and can see the corresponding spectrograms, then give their decisions in a set $\{\mathtt{Yes, No, Not~Sure}\}$ (Figure \ref{fig:zooniverseUI}). The total number of participants for this release of the dataset is 1,316. The audio clips each overlap by 1 second (50\%) in order of the $\mathtt{audio\_id}$ in the $\mathtt{csv}$ file within each group. This ensures that despite the unique $\mathtt{audio\_id}$ with its associated votes, each section of audio is covered by at least two labels.

 Figure \ref{fig:labelsperuser} shows the number of classifications made by each individual user. While the majority of participants only clicked through a few examples, the remaining users supplied votes to cover 80,101 audio segments. We represent the resulting number of labels per audio clip with a logarithmically-scaled histogram in Figure \ref{fig:labelsperclip}. The first few bins are healthily populated, with a logarithmic decline with the increase in number of labels. Of the 80,101 overlapping audio clips, 57,710 (72\%) contain more than one label, of which 10,487 (18\%) contain disagreement.

To showcase an example use of the data, we use a simple majority voting scheme in Section \ref{sec:baseline} as a baseline. We convert the ``$\mathtt{Not~Sure}$'' labels into a numerical value of 0.5, take the mean and round to the nearest integer (0 or 1). We resolve tiebreaks in favour of the mosquito class. To utilise the multiple votes that the label overlap provides, we aggregate votes from overlapping audio segments to form a new dataset of one second clips. Classification tasks using this dataset benefit from the aggregated labels per clip, resulting in higher rates of both precision and recall. We supply this processed, simplified, dataset in a separate $\mathtt{csv}$ file with the fields
 $\mathtt{\{path, subject\_set, Yes, No, Not~Sure\}}$, where $\mathtt{path}$ is the path to these new higher resolution label audio clips, and $\{\mathtt{Yes, No, Not~Sure}\}$ are columns that count the number of these occurrences per each unique one second clip.

\section{Baseline}
\label{sec:baseline}
We show an example classification use of the data in the $\mathtt{iPython}$ notebook found in the repository. 
We employ a convolutional neural network, following similar architectures used previously in mosquito detection \cite{kiskin2017mosquito,kiskin2018bioacoustic}, with two layers, with filter sizes $(3\times3)$ and unit stride length.  To deal with the data imbalance, we use a weighted cross-entropy with the class-weights, $\mathbf{w} = [1,10]$  equal to the inverse of the relative frequency of each class, calculated from the aggregated majority labels (an approximation can be viewed in Figure \ref{fig:subject_set}). Each audio recording of one second duration is transformed into the log-Mel spectrogram domain, as this feature space is commonly found to perform well in the literature \cite{gemmeke2017audio, hershey2017cnn}. Each transformed data sample is treated as a 2D image with dimensions $h \times w = 128 \times 11$. The number of rows, or height $h$, corresponds to the number of log-Mel features, and the width $w$, contains 11 samples (a result of a 0.1 second window length for the feature transform with padding). The feature-transformed data, with its corresponding one-hot-encoded majority label, is given in $\{\mathbf{X},y\} = \mathtt{\{data\_mel,label\}}$. We resolve tiebreaks in favour of the positive (mosquito) class.

We also include a copy of the data transformed to Mel-frequency cepstrum coefficients (MFCCs), as the more salient lower-dimensional representation can help algorithms such as SVMs learn models \cite{kamruzzaman2010speaker}. The attached notebook provides the core functionality such that the user may choose whichever representation is most convenient.

The data $\{\mathbf{X},y\}$ is then tenfold split with $\mathtt{sklearn}$'s $\mathtt{train\_test\_split}$ function. For reproducibility, we include the vector of random states used to partition the data in the code. The combination of majority voting, and the log-Mel features with the CNN, results in the mean confusion matrices with the standard deviations given in Figure \ref{fig:confmat}. We choose confusion matrices as metrics, as these are informative in the presence of a class imbalance. Figure \ref{fig:confmata} shows the results of training the CNN with equal class weights, $w_i =1$. This allows us to detect 39\% of the mosquito events while incurring a 2 percent false positive rate. The effect of more heavily weighting the under-represented mosquito class, with $w_1=10$, is to increase the rate of true positives at the expense of a greater number of false positives (Figure \ref{fig:confmatb}). We note, a more principled approach to deal with data imbalance can be taken using a Bayesian expected loss framework, which formally defines the desired trade-off between classes \cite{cobb2018loss}.

\section{Conclusion}
\label{sec:conclusion}
We introduce a new crowd-sourced mosquito dataset, compromised of four sources of data, containing a mixture of laboratory and field recordings. This composition of 22 hours of labelled acoustic recordings provides a realistic scenario for training machine learning techniques.
In order to demonstrate the utility of this new dataset, we offer a CNN baseline to showcase the ability to construct a mosquito detection system. We expect this acoustic dataset, alongside its metadata (such as species information and details of the recording device), to expand significantly in the coming years. We hope this will become a vital resource for those researching all aspects of malaria, and add to the existing audio datasets for bioacoustic detection and signal processing.
\section{Acknowledgements}
Paul I Howell and Dustin Miller (Centers for Disease Control and Prevention, Atlanta), Dr. Sheila Ogoma (The United States Army Medical Research Unit in Kenya (USAMRU-K)). Prof. Gay Gibson (Natural Resources Institute, University of Greenwich) and Dr. Vanessa Chen-Hussey and James Pearce at the London School of Tropical Medicine and Hygiene. For significant help and use of their field site  Prof. Theeraphap Chareonviriyaphap and members of his lab, specifically Dr. Rungarun Tisgratog and Jirod Nararak (Dept of Entomology, Kasesart University, Bangkok) and Dr. Michael J. Bangs (Public Health \& Malaria Control International SOS Kuala Kencana, Papua,  Indonesia). 
Adam D. Cobb and Ivan Kiskin are sponsored by the AIMS CDT and the EPSRC  (Grant No. EP/L015897/1). We also thank nVIDIA for the grant of a Titan Xp GPU.
\FloatBarrier

\bibliographystyle{IEEEbib}
\bibliography{strings,refs}

\begin{thebibliography}{10}

\bibitem{WHO2019}
World~Health Organization,
\newblock ``World {M}alaria {R}eport 2018,''
\newblock {\em World Health Organization International}, 2018.

\bibitem{bhatt2015effect}
Samir Bhatt, DJ~Weiss, E~Cameron, D~Bisanzio, B~Mappin, U~Dalrymple, KE~Battle,
  CL~Moyes, A~Henry, PA~Eckhoff, et~al.,
\newblock ``The effect of malaria control on {P}lasmodium falciparum in
  {A}frica between 2000 and 2015,''
\newblock {\em Nature}, vol. 526, no. 7572, pp. 207, 2015.

\bibitem{Neafsey2015}
D.E. Neafsey~et al.,
\newblock ``Mosquito genomics. {H}ighly evolvable malaria vectors: the genomes
  of 16 {A}nopheles mosquitoes.,''
\newblock {\em Science}, 2015.

\bibitem{wilkerson2015making}
Richard~C Wilkerson, Yvonne-Marie Linton, Dina~M Fonseca, Ted~R Schultz, Dana~C
  Price, and Daniel~A Strickman,
\newblock ``{M}aking {M}osquito {T}axonomy {U}seful: {A} {S}table
  {C}lassification of {T}ribe {A}edini that {B}alances {U}tility with {C}urrent
  {K}nowledge of {E}volutionary {R}elationships,''
\newblock {\em {P}lo{S} one}, vol. 10, no. 7, pp. e0133602, 2015.

\bibitem{kiskin2018bioacoustic}
Ivan Kiskin, Davide Zilli, Yunpeng Li, Marianne Sinka, Kathy Willis, and
  Stephen Roberts,
\newblock ``Bioacoustic detection with wavelet-conditioned convolutional neural
  networks,''
\newblock {\em Neural Computing and Applications: Special issue on Deep
  Learning for Music and Audio}, Aug 2018.

\bibitem{PotamitisKaggle}
Eleftherios Fanioudakis, Matthias Geismar, and Ilyas Potamitis,
\newblock ``Mosquito wingbeat analysis and classification using deep
  learning,''
\newblock pp. 2410--2414, Sept 2018.

\bibitem{hershey2017cnn}
Shawn Hershey, Sourish Chaudhuri, Daniel~PW Ellis, Jort~F Gemmeke, Aren Jansen,
  R~Channing Moore, Manoj Plakal, Devin Platt, Rif~A Saurous, Bryan Seybold,
  et~al.,
\newblock ``{CNN} {A}rchitectures for {L}arge-{S}cale {A}udio
  {C}lassification,''
\newblock in {\em 2017 {IEEE} International Conference on Acoustics, Speech and
  Signal Processing (ICASSP)}. IEEE, 2017, pp. 131--135.

\bibitem{deng2014deep}
Li~Deng, Dong Yu, et~al.,
\newblock ``Deep {L}earning: {M}ethods and {A}pplications,''
\newblock {\em Foundations and Trends{\textregistered} in Signal Processing},
  vol. 7, no. 3--4, pp. 197--387, 2014.

\bibitem{langkvist2014review}
Martin L{\"a}ngkvist, Lars Karlsson, and Amy Loutfi,
\newblock ``A review of unsupervised feature learning and deep learning for
  time-series modeling,''
\newblock {\em Pattern Recognition Letters}, vol. 42, pp. 11--24, 2014.

\bibitem{Khan1949}
Wm.~H. Offenhauser and Morton~C. Kahn,
\newblock ``The {S}ounds of {D}isease‐{C}arrying {M}osquitoes,''
\newblock {\em The Journal of the Acoustical Society of America}, vol. 21, no.
  3, pp. 259--263, 1949.

\bibitem{li2017mosquito}
Yunpeng Li, Davide Zilli, Henry Chan, Ivan Kiskin, Marianne Sinka, Stephen
  Roberts, and Kathy Willis,
\newblock ``Mosquito detection with low-cost smartphones: data acquisition for
  malaria research,''
\newblock {\em arXiv preprint arXiv:1711.06346}, 2017.

\bibitem{eLifeArticle}
Haripriya Mukundarajan, Felix Jan~Hein Hol, Erica~Araceli Castillo, Cooper
  Newby, and Manu Prakash,
\newblock ``Using mobile phones as acoustic sensors for high-throughput
  mosquito surveillance,''
\newblock {\em eLife}, vol. 6, pp. e27854, Oct 2017.

\bibitem{Simpson2014}
Robert Simpson, Kevin Page, and David De~Roure,
\newblock ``Zooniverse: {O}bserving the {W}orld's {L}argest {C}itizen {S}cience
  {P}latform,''
\newblock 04 2014, pp. 1049--1054.

\bibitem{kiskin2017mosquito}
Ivan Kiskin, Bernardo~P{\'e}rez Orozco, Theo Windebank, Davide Zilli, Marianne
  Sinka, Kathy Willis, and Stephen Roberts,
\newblock ``Mosquito {D}etection with {N}eural {N}etworks: {T}he {B}uzz of
  {D}eep {L}earning,''
\newblock {\em arXiv preprint arXiv:1705.05180}, 2017.

\bibitem{gemmeke2017audio}
Jort~F Gemmeke, Daniel~PW Ellis, Dylan Freedman, Aren Jansen, Wade Lawrence,
  R~Channing Moore, Manoj Plakal, and Marvin Ritter,
\newblock ``Audio set: {A}n ontology and human-labeled dataset for audio
  events,''
\newblock in {\em 2017 {IEEE} International Conference on Acoustics, Speech and
  Signal Processing (ICASSP)}. IEEE, 2017, pp. 776--780.

\bibitem{kamruzzaman2010speaker}
S.~M. Kamruzzaman, A.~N. M.~Rezaul Karim, Md.~Saiful Islam, and Md.~Emdadul
  Haque,
\newblock ``Speaker {I}dentification using {M}{F}{C}{C}-{D}omain {S}upport
  {V}ector {M}achine,'' 2010.

\bibitem{cobb2018loss}
Adam~D Cobb, Stephen~J Roberts, and Yarin Gal,
\newblock ``Loss-{C}alibrated {A}pproximate {I}nference in {B}ayesian {N}eural
  {N}etworks,''
\newblock {\em arXiv preprint arXiv:1805.03901}, 2018.

\end{thebibliography}

\end{document}